\documentclass[letterpaper]{article} 
\usepackage{aaai2026}  
\usepackage{times}  
\usepackage{helvet}  
\usepackage{courier}  
\usepackage[hyphens]{url}  
\usepackage{graphicx} 
\usepackage{natbib}  
\usepackage{caption} 
\usepackage{algorithm}
\usepackage{algorithmic}

\usepackage{amssymb}
\usepackage{amsmath}
\usepackage{amsthm}
\usepackage{booktabs}
\usepackage{enumitem}
\usepackage{graphicx}
\usepackage{color}
\usepackage{subfig}
\usepackage{comment}

\usepackage{mathrsfs}
\usepackage{times}
\usepackage{helvet}
\usepackage{courier}
\usepackage{xcolor}
\usepackage{newfloat}
\usepackage{listings}
\DeclareCaptionStyle{ruled}{labelfont=normalfont,labelsep=colon,strut=off} 
\floatstyle{ruled}
\newfloat{listing}{tb}{lst}{}
\floatname{listing}{Listing}
\title{Factorization-in-Loop: Proximal Fill-in Minimization for Sparse Matrix Reordering}
\author {
    Ziwei Li\textsuperscript{\rm 1,\rm 2},
    Shuzi Niu\textsuperscript{\rm 2 \thanks{Corresponding author}},
    Tao Yuan\textsuperscript{\rm 2 $^*$},
    Huiyuan Li\textsuperscript{\rm 2 $^*$},
    Wenjia Wu\textsuperscript{\rm 2}
}
\affiliations {
    \textsuperscript{\rm 1}University of the Chinese Academy of Sciences\\
    \textsuperscript{\rm 2} State Key Laboratory of Computer Science, Institute of Software Chinese Academy of Sciences\\
    liziwei2021@iscas.ac.cn, shuzi@iscas.ac.cn, yuantao@iscas.ac.cn, huiyuan@iscas.ac.cn, wuwenjia@iscas.ac.cn
}

\usepackage{bibentry}

\begin{document}

\maketitle

\begin{abstract}
Fill-ins are new nonzero elements in the summation of the upper and lower triangular factors generated during LU factorization. For large sparse matrices, they will increase the memory usage and computational time, and be reduced through proper row or column arrangement, namely matrix reordering. Finding a row or column permutation with the minimal fill-ins is NP-hard, and surrogate objectives are designed to derive fill-in reduction permutations or learn a reordering function. However, there is no theoretical guarantee between the golden criterion and these surrogate objectives. Here we propose to learn a reordering network by minimizing \(l_1\) norm of triangular factors of the reordered matrix to approximate the exact number of fill-ins. The reordering network utilizes a graph encoder to predict row or column node scores. For inference, it is easy and fast to derive the permutation from sorting algorithms for matrices. For gradient based optimization, there is a large gap between the predicted node scores and resultant triangular factors in the optimization objective. To bridge the gap, we first design two reparameterization techniques to obtain the permutation matrix from node scores. The matrix is reordered by multiplying the permutation matrix. Then we introduce the factorization process into the objective function to arrive at target triangular factors. The overall objective function is optimized with the alternating direction method of multipliers and proximal gradient descent. Experimental results on benchmark sparse matrix collection SuiteSparse show the fill-in number and LU factorization time reduction of our proposed method is 20\% and 17.8\% compared with state-of-the-art baselines.
\end{abstract}

\begin{links}
    \link{Code}{https://github.com/plumvvvv/PFM}
\end{links}

\section{Introduction}
Many large sparse systems in scientific applications, such as computational fluid dynamics and materials science, are often reduced to linear equations like \(Ax=b\). $A\in \mathbb{R}^{n \times n}$ is a sparse matrix, defined as a matrix predominantly comprised of zero entries~\cite{Wilkinson1971}. $x$ and $b$ are the solution and right-hand side vector or matrix respectively. Direct solvers are commonly employed to solve this system due to their ability to provide exact solutions. The core idea behind these solvers is to factorize the matrix \(A\) into its lower triangular factor \(L\) and upper triangular factor \(U\), such that \(A = L \times U\), where \(L\) has ones on the diagonal and \(U\) is an upper triangular matrix. The solution \(x\) is efficiently computed through two simpler linear equations \(Ly = b\) and \(Ux = y\). The number of nonzero elements in both $L$ and $U$ factors determines the memory usage and computational cost of the whole solving process. 

Actually the number of nonzero elements in both factors is usually larger than that in the original matrix due to the factorization process. Those new nonzero elements are referred to as fill-ins and their generation is intuitively described in the Gaussian elimination process, involving transformation and results structured as $L$ and $U$ factors. Each row elimination step involves multiplying the pivot row by a coefficient and adding it to the other rows that have not yet been eliminated, which may introduce new nonzero elements in positions that were previously zero in the original matrix, i.e. fill-ins. Different row or elimination orderings, i.e. row or column permutations, usually lead to different fill-ins. How to determine the row or column permutations with minimal fill-ins lies at the heart of direct solvers in terms of memory and computation efficiency. 

Due to its NP-hard nature~\citep{Yannakakis1981}, most matrix reordering algorithms utilize their adjacency graphs to take advantage of sparse structures and derive graph-theoretical criteria to eliminate nodes. Some algorithms~\citep{AMD,MMD} sequentially select nodes with the minimal degree to eliminate accounting for its neighbors. Other algorithms~\citep{RCM,CM} assume that the reordered matrix should satisfy some desired properties, such as the concentration on the major diagonal. Another class of algorithms~\citep{spectral,ND} focus on how to identify these relatively dense rows or columns. Moreover, the criteria are reordering functions in the form of deep neural networks, whose parameters are optimized with deep learning techniques. AlphaElim~\cite{alphaelim} is such an example to optimize reordering network parameters in a reinforcement learning framework due to the non-differentiable nature of the fill-in number. Due to its sequential inference process, it fails to scale to large matrices.
As far as we know, the theoretical relationship between these criteria and the fill-in number is unclear though empirical results show their feasibility in some matrices.

\begin{figure}[!htbp]
    \centering
    \includegraphics[width=0.7\linewidth]{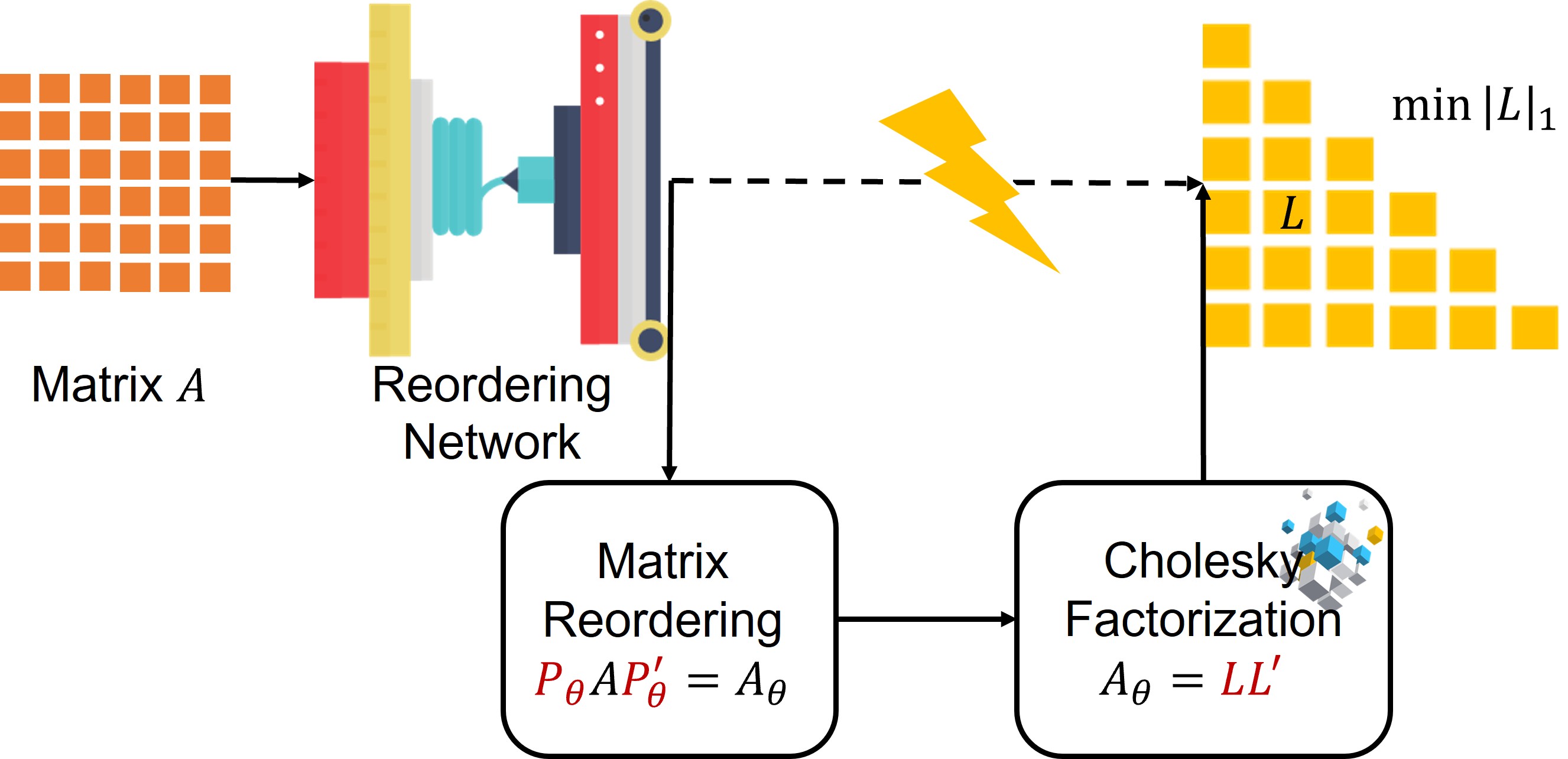}
    \caption{Illustration of introducing Cholesky factorization process into the network optimization process.}
    \label{fig:illustration}
\end{figure}

Here we propose to minimize the \(l_1\) norm of triangular factors to learn a reordering network, which is a well-known convex surrogate of the precise fill-in number, \(l_0\) norm of triangular factors. 
For inference, we design the reordering network with a graph neural network backbone to predict node scores like the left part of Figure~\ref{fig:illustration}, according to which matrix rows or columns are reordered.
For gradient based optimization, we first transform the score function into a differentiable permutation matrix with reparameterization techniques~\cite{SoftRank}, perform matrix reordering through matrix multiplication operations like the left additional module of Figure~\ref{fig:illustration}. Then we introduce Cholesky factorization of the reordered matrix into the training process in the right additional module of Figure~\ref{fig:illustration} and can achieve the goal of optimizing the \(l_1\) norm of the Cholesky factor. In all, we minimize \(l_1\) norm of the factor constrained by the Cholesky factorization equation.

Specifically we solve this constrained optimization problem with the alternating direction method of multipliers~\cite{admm}, decomposing it into the factor and permutation matrix related optimization subproblems. For the factor optimization, it minimizes a factorization loss with \(l_1\) norm by proximal gradient descent~\cite{boyd2004convex}. For the permutation matrix optimization, it minimizes a factorization loss with Adam optimizer. The whole optimization framework is referred to as Proximal Fill-in Minimization(PFM). We conduct experiments on benchmark datasets SuiteSparse~\cite{suitesparse}, and results show the superiority of our proposed method PFM in terms of the nonzero element number and LU factorization time. Our major contributions include \(l_1\) norm minimization of the Cholesky factor to approximate its fill-in number, differentiable matrix reordering layer to rearrange rows and columns according to the predicted scores, and the Cholesky factorization enhanced loss for optimization.

\section{Related Work}
Sparse matrix reordering has traditionally been dominated by graph-theoretic methods. 
Given a sparse symmetric matrix $A\in R^{n\times n}$, its adjacency graph is denoted as $G=(V,E)$, \(V\) and \(E\) are the node and edge set of $G$ respectively, where $e_{ij}\in E$ means $a_{ij}\neq 0$. Here we only focus on LU factorization of symmetric matrices, i.e. Cholesky factorization. Graph theoretical reordering algorithms aim at obtaining a node ordering based on some designed criteria from graph structure $G$. With the aid of deep learning techniques, those reordering scores are supposed to be predicted by deep neural networks \(f_{\theta}\), whose parameters \(\theta\) are learned by minimizing the fill-in number generated in the subsequent Cholesky factorization, referred to as learning matrix reordering. 
In this section, we provide a detailed overview of both traditional graph-theoretic algorithms and emerging deep learning methods.

Graph-theoretic approaches leverage the reordering criterion from graph structures. 
Some algorithms design reordering scores mainly based on node degrees. Minimum Degree (MD) algorithm proposed by ~\citet{MD} selects the node with the smallest degree at each elimination step. Multiple Minimum Degree (MMD)~\citep{MMD} selects multiple nodes at once. Approximate Minimum Degree (AMD)~\citep{AMD} simplifies degree calculations for better performance. Other methods utilize matrix bandwidth related scores. Cuthill-McKee (CM)~\citep{CM} algorithm and its reverse variant, Reverse Cuthill-McKee (RCM)~\citep{RCM}, are widely used to reduce matrix bandwidth. CM algorithm first selects a pseudo-peripheral node, then performs a breadth-first search on the adjacency graph, and finally sorts nodes by ascending degree. Similarly, spectral ordering methods exploit the spectral properties of the graph Laplacian to reduce envelope size, such as the Fiedler vector method~\citep{spectral}. However, computing the Fiedler vector is computationally expensive especially for large graphs.  

Graph partitioning based reordering methods attempt to first partition the graph by minimizing the connections between subgraphs while maximizing the connections within each subgraph. Then they tend to order node groups and refine the node ordering in each group, such as Nested Dissection~\cite{ND} and multilevel \(k\)-way partitioning~\citep{kway}. METIS~\citep{metis} and SCOTCH~\citep{scotch} efficiently implement both ND and multilevel partitioning, producing effective elimination orderings. Graph-theoretic methods are intuitively designed and there is no theoretical guarantee of its approximation ratio to the precise fill-in number minimization. 
\begin{table}[!htbp]
    \centering
    \small
    \begin{tabular}{c|c|c}
    \hline
       algorithm  & worst case time complexity & parallizability \\
       \hline
       AMD  & \(\mathcal{O}(|E|\cdot|V|)\) & low\\
       Metis & \(\mathcal{O}(|E|\cdot\log|V|)\) & low\\
       Spectral & \(\mathcal{O}(|V|^3)\) & low\\
       AlphaElim & \(\mathcal{O}(|V|\cdot \text{CNN}_{\gamma})\) & low\\
       UDNO & \(\mathcal{O}(\text{GNN}_{\phi})\) & high\\
       PFM & \(\mathcal{O}(\text{GNN}_{\phi})\) & high\\
       \hline
    \end{tabular}
    \caption{Time complexity analysis of different reordering methods. 
        \(\text{CNN}_{\gamma}\) and \(\text{GNN}_{\phi}\) mean the time complexity of convolutional and graph neural networks with parameter \(\gamma\) and \(\phi\) respectively. }
    \label{tab:timecomplexity}
\end{table}

Recently, deep learning methods show promising results in numerical solvers~\cite{LearnfromLA}. Neural Incomplete Factorization method~\cite{learnincomplete} uses graph neural networks (GNNs) to learn incomplete LU (ILU) factorization preconditioners, significantly reducing GMRES iteration counts and improving spectral properties for large sparse linear systems. An end-to-end graph neural network (GNN) framework~\cite{LearnfromLA} is proposed as preconditioners for conjugate gradient (CG) solvers by representing sparse matrices as graphs and minimizing the spectral radius of preconditioned systems, significantly accelerating iterative solution convergence. 

For matrix reordering in direct solvers, deep neural networks usually act as a policy in a reinforcement learning framework for fill-in optimization. The policy network in \citet{GP} is to select small separators in a graph partitioning manner. Convolutional neural network based policy in \cite{alphaelim} is to select rows or columns to be eliminated to minimize the fill-ins from Gaussian elimination. Besides, ~\citet{Acceleration} uses graph neural networks to predict the potential fill-ins generated from the LU factorization of the current matrix without de facto factorization, and makes the choice of reordering algorithms convenient. The major drawback of existing deep reinforcement learning based reordering method lies in the sequential selection process, facing the high time complexity in the inference process in Table~\ref{tab:timecomplexity}. UDNO~\cite{udno} is an exception that uses multi-grid GNN to predict a node reordering all at once, minimizing an expected envelope-like loss without theoretical guarantee.

\section{Proximal Fill-in Minimization Framework}
\begin{figure}[!htbp]
    \centering
    \includegraphics[width=\linewidth]{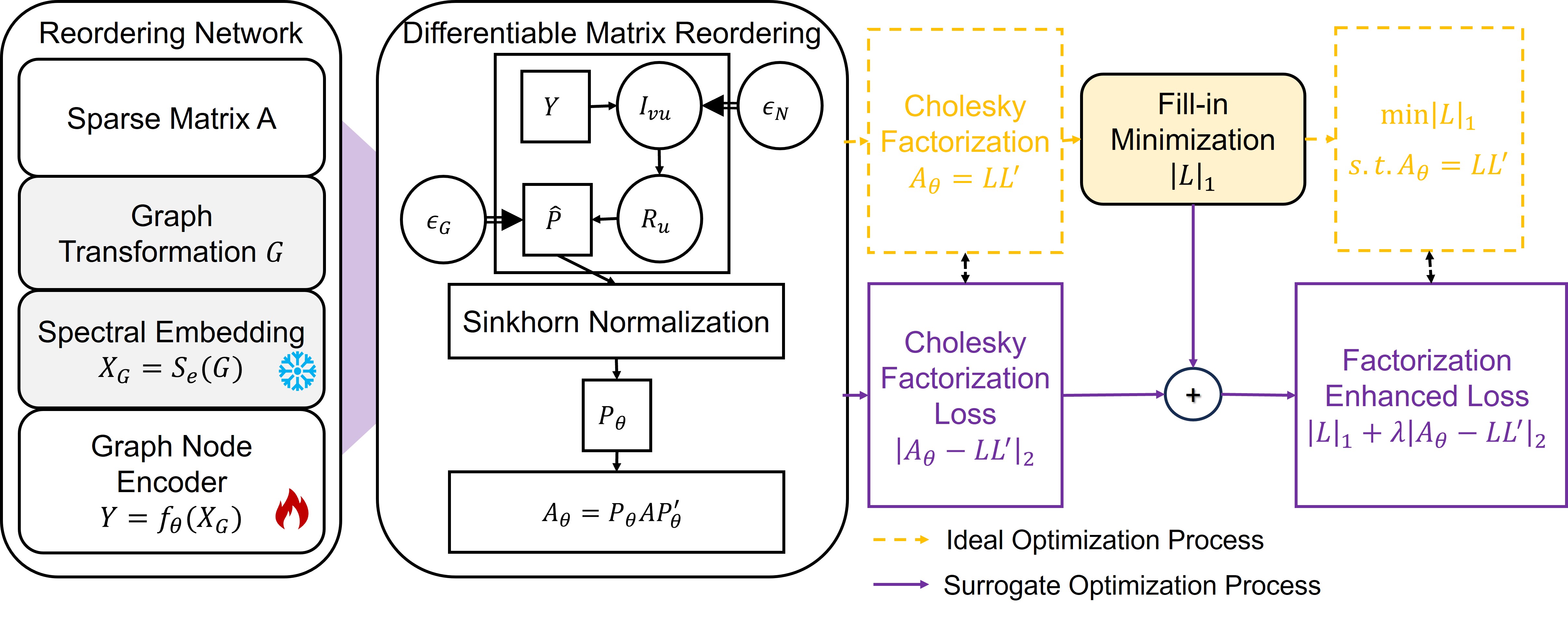}
    \caption{Proximal Fill-in Minimization Framework}
    \label{fig:arch}
\end{figure}

To minimize the \(l_1\) norm of triangular factors as fill-in surrogate, we introduce a Cholesky factorization process into the training loop, i.e. Factorization-in-Loop. \(l_1\) norm of triangular factors is a convex approximation to the number of fill-ins. In the inference process of Figure~\ref{fig:arch}, each input sparse symmetric matrix \(A\) is sequentially fed into graph transformation layer, spectral embedding layer and graph node encoder. The reordering network finally predicts all node scores. In the training process, an additional layer, i.e. differentiable matrix reordering layer, is introduced to derive a reordered matrix for gradient computation in the backward process. Finally the factorization error loss is introduced to the \(l_1\) norm of triangular factors as the optimization objective to simulate the constrained optimization problem in the yellow box of Figure~\ref{fig:arch}. We choose proximal gradient descent method to optimize the surrogate fill-in number, i.e. \(l_1\) norm, and the whole framework is referred to as Proximal Fill-in Minimization.

\subsection{Convex Approximation of the Fill-in Number}
Fill-ins refer to the new nonzero elements in the Cholesky factor \(L\) compared with those in the original matrix \(A\). The number of fill-ins in factor \(L\) is ideal but hard to optimize. Minimizing the fill-in number equals to optimizing the nonzero number of factor \(L\), formalized as entrywise \(l_0\) norm \(\|L\|_0\). Theoretically, it has been proven that \(\|L\|_1\) is a convex relaxation of \(\|L\|_0\)~\cite{convexanalysis}. Thus \(\|L\|_1\) is also a convex approximation of the fill-in number in \(L\). \(l_1\) norm of the factor \(\|L\|_1\) is computed as the sum of the absolute values of elements in factor \(L\) like Eq.(\ref{eq:l1norm}). Practically, \(l_1\) norm minimization is often used for its computational efficiency and ability to produce sparse solutions. The minimization of \(\|L\|_1\) encourages the factor matrix $L$ to be as sparse as possible. \(l_1\) norm serves as a computationally feasible and theoretically guaranteed alternative in sparse optimization problems.
\begin{equation}
||L||_1=\sum_{i=0}^{n-1}\sum_{j=0}^{n-1}|L_{ij}|
    \label{eq:l1norm}
\end{equation}

\subsection{Reordering Network}
Given a sparse symmetric matrix \(A\in\mathbb{R}^{n\times n}\), the reordering network first transforms it into a graph format \(G=(V,E)\) through graph transformation layer. As we mentioned before, each node in \(G\) corresponds to a row or column in \(A\), and each edge corresponds to a nonzero element in \(A\). Node feature $X\in\mathbb{R}^{n\times 1}$ is initialized randomly in Eq.(\ref{eq:initx}). With this initialization, the spectral embedding module estimates the Fiedler vector, i.e. the second smallest eigenvector of the Laplacian, through a multi-grid graph neural network~\citep{spectralembedding}. As we know, exact spectral information is usually expensive to obtain. So here we use its pretrained weights to estimate the spectral embedding. Moreover, the effectiveness of this pretrained model \(S_e\) has been verified in graph partitioning task. The spectral embedding of nodes are updated as Eq.(\ref{eq:se}).
\begin{equation}
X=\text{randn}(n)
    \label{eq:initx}
\end{equation}
\begin{equation}
X_G=S_e(X)
    \label{eq:se}
\end{equation}

To refine the task specific information from the current node embedding \(X_G\), we utilize a graph node encoder for matrix reordering. Only parameters \(\theta\) in this encoder are updated during the following training process. \(G\) is sparse with only about 10\% edges and it is difficult to capture useful structure information from this weakly connected graph. At the same time, it poses a great challenge to graph node encoder. We employ graph neural networks better at capturing multi-scale structural information as graph node encoder, such as multi-grid graph neural network~\citep{spectralembedding} and GraphUnet~\cite{graphunet} in our experiments. Through this node encoder, the reordering network \(f_{\theta}\) predicts the reordering scores for all nodes in \(V\) as Eq.(\ref{eq:nodescore}). 
\begin{equation}
Y=f_{\theta}(X_G)
    \label{eq:nodescore}
\end{equation}

 \begin{figure}[!htbp]
 \centering
 \subfloat[Score Function]{
 \includegraphics[width=0.2\textwidth]{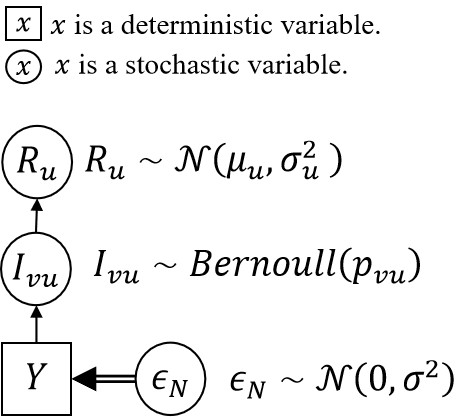}
 }
 \quad
 \subfloat[Rank Distribution]{
 \includegraphics[width=0.17\textwidth]{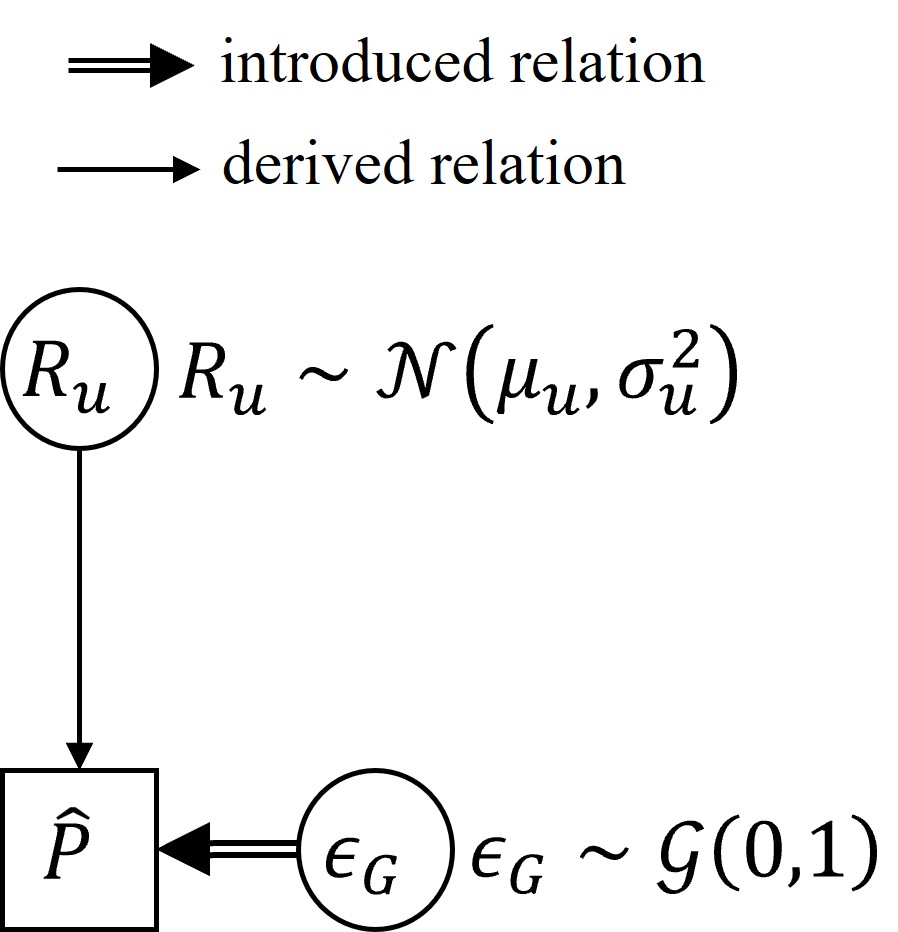}
 }
 \caption{Score and Rank Distribution Reparameterization.}
 \label{fig:reparam}
 \end{figure}
\subsection{Differentiable Matrix Reordering Layer}
For inference, nodes in \(V\) are sorted according to scores \(Y\). However, these sorting operations cannot be applied to training phase due to its non-differentiability. Only for optimization, we design two reparameterization techniques to obtain a differentiable permutation matrix \(P_{\theta}\) considering the inference efficiency. Specifically, the ranking scores \(Y\) are transformed into a Gaussian rank distribution matrix \(\hat{P}\), which is subsequently reparameterized into a permutation matrix \(P_{\theta}\) in Figure~\ref{fig:reparam}. The matrix reordering is performed in a differentiable way like Eq.(\ref{eq:matrixreordering}), namely differentiable matrix reordering layer.
\begin{equation}
A_{\theta}=P_{\theta}AP'_{\theta}
    \label{eq:matrixreordering}
\end{equation}

With the \textbf{first reparameterization} technique like Figure~\ref{fig:reparam} (a), we add the Gaussian noise \(\epsilon_{N}\) to score $Y$, where \(\epsilon_N\sim\mathcal{N}(0,\sigma^2)\). For each node \(u\), its score in \(Y\), denoted as \(Y_u\), becomes a Gaussian random variable \(\mathcal{Y}_u\), following a Gaussian distribution \(\mathcal{N}(Y_u,\sigma^2)\). The comparison result between node \(u\) and \(v\) is a Bernoulli random variable \(I_{vu}\). Its probability distribution \(P(I_{vu}=1)\) is denoted as \(p_{vu}\) for short meaning the probability that node $v$ is ranked above \(u\). Its specific form is derived from their node score distribution as Eq.(\ref{eq:pvu}).
\begin{equation}
  p_{vu}=\text{Pr}(\mathcal{Y}_v-\mathcal{Y}_u>0)=\int_0^{\infty}\mathcal{N}(Y_v-Y_u,2\sigma^2)
  \label{eq:pvu}
\end{equation}

According to the sorting algorithm, the position/rank of node \(u\) in the ordered list, denoted as \(R_u\), is determined by its comparison results with the other nodes in \(V\) like Eq.(\ref{eq:rank}). Suppose \(I_{vu}\) are independent random variables. When \(n=|V|\) is sufficiently large, the rank distribution of node \(u\) is supposed to follow a normal distribution $R_u\sim \mathcal{N}(\mu_u,\sigma_u^2)$. Its mean and variance can be derived according to the rules for sum of independent Gaussian random variables as Eq.(\ref{eq:meanvar}). The rank distribution matrix \(\hat{P}\) is constructed from node rank distributions, where each element is defined as Eq.(\ref{eq:rankdistribution}). It satisfies that the row sum is nearly 1, but the permutation matrix requires a square binary matrix with exactly one entry of 1 in each row/column. This rank distribution matrix is a good prototype permutation matrix, which is fed into the following Gumbel-Sinkhorn process.
\begin{equation}
R_u=\sum_{v\in V-{u}}I_{vu}\sim \mathcal{N}(\mu_u,\sigma_u^2)
    \label{eq:rank}
\end{equation}
\begin{equation}
\mu_u=\sum_{v\in V-{u}}p_{vu},\sigma_u^2=\sum_{v\in V-{u}}p_{vu}(1-p_{vu})
    \label{eq:meanvar}
\end{equation}
\begin{equation}
\hat{P}(u,i) = \text{Pr}(i - 0.5 < R_u < i + 0.5)
    \label{eq:rankdistribution}
\end{equation}

Given the rank distribution matrix $\hat{P}$, we employ the \textbf{second reparameterization} technique to obtain an approximate permutation matrix illustrated in Figure~\ref{fig:reparam}(b). Some differentiable sorting layers have been proposed such as Gumbel Sinkhorn~\citep{gumbelsinkhorn} and NeuralSort~\citep{neuralsort}. We adopt Gumbel Sinkhorn method~\footnote{\url{https://github.com/HeddaCohenIndelman/Learning-Gumbel-Sinkhorn-Permutations-w-Pytorch}} to avoid additional learnable parameters and infer efficiently. According to the Sinkhorn–Knopp theorem~\citep{sinkhorntheorem}, any positive matrix that is sufficiently well-conditioned can be transformed into a doubly stochastic matrix through alternating row and column normalizations. This property enables the Gumbel Sinkhorn algorithm to approximate permutation matrices without requiring explicit sorting, while maintaining its differentiability. 

Gumbel Sinkhorn process is mainly composed of two steps. First, it adds Gumbel noise \(\epsilon_G\) to each element of the input rank distribution matrix as Eq.(\ref{eq:addgumbelnoise}). This step introduces stochasticity into the permutation process, enabling the model to explore different rankings while remaining fully differentiable. To improve numerical stability, we bring the computation into the log space. Second, the resulting perturbed log-probability matrix logP is refined through Sinkhorn normalization iterations, i.e., alternating row-wise and column-wise softmax normalization. This process enforces the doubly stochastic constraint. The final approximate permutation matrix \(P_{\theta}\) is obtained by \(\exp(\text{logP})\), with each row and column containing a single entry close to one and the rest close to zero.
\begin{equation}
\text{logP}=\log\hat{P}+\epsilon_G
    \label{eq:addgumbelnoise}
\end{equation}
 
\subsection{Factorization Enhanced Loss Function}
We obtain a new matrix \(A_{\theta}\) with reordered rows and columns according to the derived permutation matrix \(P_{\theta}\) as Eq.(\ref{eq:matrixreordering}). Starting with the reordered matrix \(A_{\theta}\), we still need a Cholesky factorization step to obtain the triangular factor \(L\). Naturally, we introduce the factorization step as a constraint. The constrained optimization problem is to minimize the entrywise \( l_1 \) norm of the factor matrix \( L \), subject to the constraint \( A_{\theta} = LL' \) as Eq.(\ref{eq:constrainedproblem}).
\begin{equation}
\begin{aligned}
    \min&\quad \|L\|_1\\
    s.t.&\quad P_{\theta} A P_{\theta}' = L L'
\end{aligned}
    \label{eq:constrainedproblem}
\end{equation}

To realize an end-to-end joint optimization framework, we incorporate the factorization constraint into the final loss function in the augmented Lagrangian form as Eq.(\ref{eq:loss}). \(\Gamma\in\mathbb{R}^{n\times n}\) is the Lagrange multiplier and also called dual variable. \(\rho\) is the penalty parameter controlling constraint satisfaction and we set it to 1 in our experiments. The whole loss in Eq.(\ref{eq:loss}) is referred to as Factorization Enhanced Loss Function. This loss function can be optimized by Alternating Direction Method of Multipliers (ADMM)~\cite{admm}. Generally, we add a factorization constraint to the original \(l_1\) norm of the factor \(\|L\|_1\) from the loss function perspective. 
\begin{equation}
\begin{aligned}
\mathcal{L}_\rho(L,P_{\theta},\Gamma) &= \|L\|_1 \\
&+ \underbrace{trace(\Gamma^T (P_{\theta}AP'_{\theta}-LL'))}_{\text{dual term}} \\
&+ \frac{\rho}{2} \| P_{\theta}AP'_{\theta}-LL' \|_2^2
\end{aligned}
    \label{eq:loss}
\end{equation}

\subsection{Optimization Algorithm}
From the optimization algorithm perspective, we introduce an incomplete Cholesky factorization process into the optimization process. There are three kinds of variables in the factorization enhanced loss in Eq.(\ref{eq:loss}) including \(L\), \(\theta\) and dual variable \(\Gamma\). The minimization of this loss function is decomposed into the following three optimization subproblems for these three kinds of variables. For each training iteration \(k\) in ADMM, we need to update the following three kinds of variables as Eq.(\ref{eq:updateadmm}).
\begin{equation}
    \begin{aligned}
    L^{k+1} &= \arg\min_L \mathcal{L}_\rho(L,P_{\theta}^k,\Gamma^k) \\
\theta^{k+1} &= \arg\min_{\theta} \mathcal{L}_\rho(L^{k+1}, P_{\theta}, \Gamma^k)-\|L^{k+1}\|_1 \\
\Gamma^{k+1} &= \Gamma^k + \rho (P_{\theta}^{k+1}AP_{\theta}^{k+1\,\prime}-L^{k+1}L^{k+1\,\prime})
    \end{aligned}
    \label{eq:updateadmm}
\end{equation}

To optimize the \(L\)-related subproblem, the objective function involves all three terms in Eq.(\ref{eq:loss}). The optimization of \(L\) is decomposed into gradient step and proximal step. (1) Gradient step: The gradient of the dual term and \(l_2\) term of \(\mathcal{L}_{\rho}\) is easy to compute, denoted as \(\nabla_{L}\mathcal{L}_{\rho,\text{dual}+l_2}\). \(L\) is updated according to this gradient in Line 10 of Algorithm~\ref{alg:fillinmin}. (2) Proximal step: For \(l_1\) norm minimization, the subgradient descent method is often adopted, but it fails to explicitly encourage sparsity and converge slowly. Here we adopt the proximal gradient descent for \(\|L\|_1\) optimization. The proximal gradient for \(\|L\|_1\) is computed as Eq.(\ref{eq:proxgrad}). This operation shrinks each element \(L(i,j)\) by \(\eta\), ensuring that small values are driven to zero, which directly corresponds to the sparsity-promoting behavior of the \(l_1\) norm.
Generally, \(l_2\) norm term is the loss function adopted by learning incomplete LU factorization for a certain fill-in level to obtain an approximate preconditioner for better convergence of iterative solvers. In this sense, we incorporate the incomplete Cholesky factorization process into the optimization process.
\begin{equation}
\mathcal{S}_{\eta}(L(i,j)) = \text{sign}(L(i,j)) \cdot \max\left( |L(i,j)| - \eta, 0 \right)
\label{eq:proxgrad}
\end{equation}

\begin{algorithm}[!htbp]
\caption{Proximal Fill-in Minimization Method}
\small
\label{alg:fillinmin}
    \begin{algorithmic}[1]
    \REQUIRE Training Matrix Set $\{A\}$;
    \ENSURE Network parameters $\theta$;
        \FOR {$m=1\ldots M$}
          \FOR {$A\in\{A\}$}
            \STATE Transform $A$ into its graph $G$;
            \STATE $Y = f_{\theta}(S_e(G))$;
            \STATE $P_{\theta}=\text{GumbelSinkhorn}(Y\rightarrow I_{vu}\rightarrow R_u\rightarrow\hat{P})$; 
            \STATE Initialize $L=\text{tril}(\text{randn}(A.\text{shape}))$;
            \STATE Initialize $\Gamma=\text{randn}(A.\text{shape})$;
            \FOR {$k=1\ldots n_{\text{ADMM}}$}
            \STATE  {\%$L$-update: gradient step}
            \STATE $L=L-\eta\nabla_{L}\mathcal{L}_{\rho,\text{dual}+l_2}$;
            \STATE {\% $L$-update: proximal operator step}
            \STATE Compute $\mathcal{S}_{\eta}(L)$ with each entry in Eq.(\ref{eq:proxgrad});
            \STATE $L=\text{tril}(\mathcal{S}_{\eta}(L))$;
            \STATE {\% $\theta$-update }
            \STATE Update $\theta$ through Adam in Eq.(\ref{eq:updateadmm});
            \STATE $Y = f_{\theta}(S_e(G))$;
            \STATE $P_{\theta}=\text{GumbelSinkhorn}(Y\rightarrow \cdots\rightarrow\hat{P})$;
            \STATE {\% $\Gamma$-update }
            \STATE $\Gamma = \Gamma + \rho (P_{\theta}AP_{\theta}'-LL')$
            \ENDFOR
          \ENDFOR
        \ENDFOR
        \RETURN $\theta$
    \end{algorithmic}
\end{algorithm}

The training loop for minimizing \(\mathcal{L}_\rho(L,P_{\theta},\Gamma)\) is shown in Algorithm~\ref{alg:fillinmin}. The basic idea is the inner loop Line \(8\sim20\) with ADMM. It is mainly composed of three optimization steps. Line \(9\sim10\) is to update \(L\) with the gradient step, simulating the incomplete Cholesky factorization process. Line \(11\sim12\) is to compute the proximal operator of the factor \(L\) and Line 13 keeps only the lower triangular part of \(L\). Line \(14\sim15\) is to update \(\theta\) with the current \(L\). This indicates that the sparser factor \(L\) benefits the better gradient of \(\theta\) towards fill-in reduction direction, which lies at the heart of our proposed method. So the latest scores \(Y\) are predicted in Line 16 with the updated \(\theta\) and the corresponding permutation matrix is also recalculated in Line 17. The final optimization step from Line 18 to 19 is to update dual variable \(\Gamma\) to satisfy the factorization constraint. The intermediate loop from Line 2 to 21 learns each matrix in the whole training set as an epoch, which will be repeated $M$ epochs in the outer loop.

\begin{table*}[!htbp]
\centering

\footnotesize
\begin{tabular}{|c|p{0.7cm}p{0.7cm}p{0.7cm}p{0.7cm}p{0.7cm}p{0.7cm}p{0.7cm}|
p{0.7cm}p{0.7cm}p{0.7cm}p{0.7cm}p{0.7cm}p{0.7cm}p{0.7cm}|}
\hline
& \multicolumn{7}{c|}{Fill-in Ratio} & \multicolumn{7}{c|}{LU factorization Time (second)}\\
  \hline
   & CFD	&MRP	&SP	&2D3D	&TP	&Other	&All
& CFD	&MRP	&SP	&2D3D	&TP	&Other	&All\\

 \hline
Natural 
&361.23 	&83.46 	&185.60 	&474.71 	&523.32 	&580.21 	&361.73 	
&466.81 	&72.81 	&913.91 	&734.79 	&873.11 	&745.10 	&679.09 \\
\hline
AMD
&386.75 	&402.85 	&194.35 	&544.43 	&642.35 	&360.49 	&344.55 	
&372.75 	&707.41 	&343.53 	&560.56 	&2960.28 	&477.03 	&535.30 \\
Metis
&73.49 	&59.37 	&56.02 	&85.62 	&93.75 	&\textbf{146.68} 	&91.19 	
&46.39 	&49.68 	&245.34 	&35.93 	&45.32 	&105.93 	&123.52 \\
Fiedler
&\textbf{56.42} 	&\textbf{46.28} 	&48.39 	&78.23 	&69.59 	&157.97 	&86.71 	
&20.24 	&41.34 	&74.78 	&19.49 	&\textbf{11.26} 	&106.80 	&65.28 \\
\hline
\(S_e\)
&61.62 	&52.09 	&55.66 	&79.95 	&72.41 	&157.95 	&90.61 	
&17.14 	&63.49 	&80.63 	&14.04 	&13.90 	&118.96 	&72.31 \\
GPCE
&61.61 	&47.89 	&49.98 	&80.54 	&70.33 	&164.64 	&90.52 	
&20.77 	&54.75 	&72.83 	&11.79 	&15.35 	&105.05 	&65.21 \\
UDNO
&59.57 	&48.25 	&46.67 	&79.75 	&75.41 	&154.82 	&86.29 	
&\textbf{16.36} 	&47.22 	&51.92 	&12.48 	&14.09 	&105.22 	&57.49 \\
PFM
&58.30 	&48.16 	&\textbf{45.32} 	&\textbf{77.34} 	&\textbf{69.28} 	&157.65 	&\textbf{86.14} 	
&17.59 	&\textbf{26.40} 	&\textbf{43.35} 	&\textbf{11.68} 	&16.73 	&\textbf{91.94} 	&\textbf{48.80} \\
\hline
\end{tabular}
\caption{Performance comparison across various ordering methods on benchmark test set SuiteSparse.}\label{tab:effect}
\end{table*}
\section{Experiments}
Here we evaluate the feasibility of reordering methods in terms of both the fill-ins and their running time. Comprehensive experiments on benchmark sparse matrix collections are conducted to verify both the feasibility and scalability of the proposed method.

\subsection{Experimental Setting}
We conduct experiments on a well designed training set proposed by~\citet{spectralembedding}. It consists of three kinds of sparse matrices:  (1) 2D/3D matrix subset from SuiteSparse; (2) Matrices generated by Delaunay method within GradeL, Hole3, and Hole6 geometries; (3) Matrices generated by finite element method within above geometries. Sparsity patterns of these training matrices vary much. Pretraining the spectral embedding module \(S_e\) needs such \(5000\) matrices with size ranging from 100 to 5000. Parameters of \(S_e\) are frozen during the proximal fill-in minimization process. This process utilizes \(100\) matrices of such three kinds with size ranging from 100 to 500.

The test set is a subset of SuiteSparse matrix collection, public benchmark test set of real-world sparse matrices from scientific computing and engineering applications~\citep{suitesparse}, adopted by UDNO~\cite{udno}. It covers most applications in SuiteSparse collection and contains 44, 25, 16, 12, 5, 46 matrices for Structural problem (SP), Computational Fluid Dynamics Problem (CFD), Model Reduction problem (MRP), 2D and 3D discretized problem (2D3D), Thermal problem (TP) and other problems, respectively. The total number of test matrices is 148. The matrix size ranges from $10,000$ to $1,000,000$. There is no intersection between training and testing 2D3D matrices.

The evaluation pipeline is performed as follows. Given a matrix $A$, each method computes a permutation matrix $P_*$. The reordered matrix $A_*$ is obtained as $P_* A P_*'$. It is fed into Python interface $splu$ of SuperLU to obtain LU factorization $A_*=L_*U_*$. The number of fill-in generated during this factorization is measured as $\text{nnz}(L_*)+\text{nnz}(U_*)-\text{nnz}(A)$. nnz(\(\cdot\)) means the number of nonzero elements in the matrix. Ignoring the matrix size influence, we normalize the number of fill-ins as fill-in ratio as follows:
\begin{equation}
    \frac{\text{nnz}(L_*)+\text{nnz}(U_*)-\text{nnz}(A)}{\text{nnz}(A)}.
\end{equation}
Fill-ins are used to roughly measure the computation and memory cost of LU factorization. The running time for LU factorization is the other important metric. The start and end timestamps for $splu$ are denoted as $t_{begin}$ and $t_{end}$, and the execution time for LU factorization is  $t_{end}-t_{begin}$ measured in seconds on the GTX4090 GPU.

There are two kinds of baselines: (1) Graph theoretical methods include AMD~\citep{AMD}, Metis~\citep{metis}, and Fiedler Vector~\citep{spectral}; (2) Deep Learning methods contain the spectral embedding model \(S_e\)~\cite{spectralembedding}, GPCE and UDNO~\cite{udno}. Taking the spectral embedding as input, GPCE is composed of two SAGEConv layers, whose parameters are also trained on the same set as Proximal Fill-in Minimization method PFM. It utilizes Pairwise Cross Entropy loss to minimize the discrepancy between the predicted ordering and the approximate ground truth, i.e. the minimal fill-in ordering among AMD, Metis and Fiedler. Natural method means no matrix reordering before factorization.

Hyperparameters are set as follows. Learning rates ranging from $10^{-5}$ to $0.1$ are chosen with the \(l_1\) norm on the training set. The final setting is 0.01 for all the optimization steps in the algorithm. \(\sigma\) in the differentiable matrix reordering layer is set to $0.001$. \(\rho\) in the factorization enhanced loss is set to $1$. Graph node encoder in PFM reordering network is chosen among Multi-grid GNN~\cite{spectralembedding} and GraphUnet~\cite{graphunet} in the ablation study. We mainly use Multi-grid GNN in the experiments. 

\begin{figure*}[!htbp]
\centering
\subfloat[Fill-in Ratio]{
\includegraphics[width=0.33\textwidth]{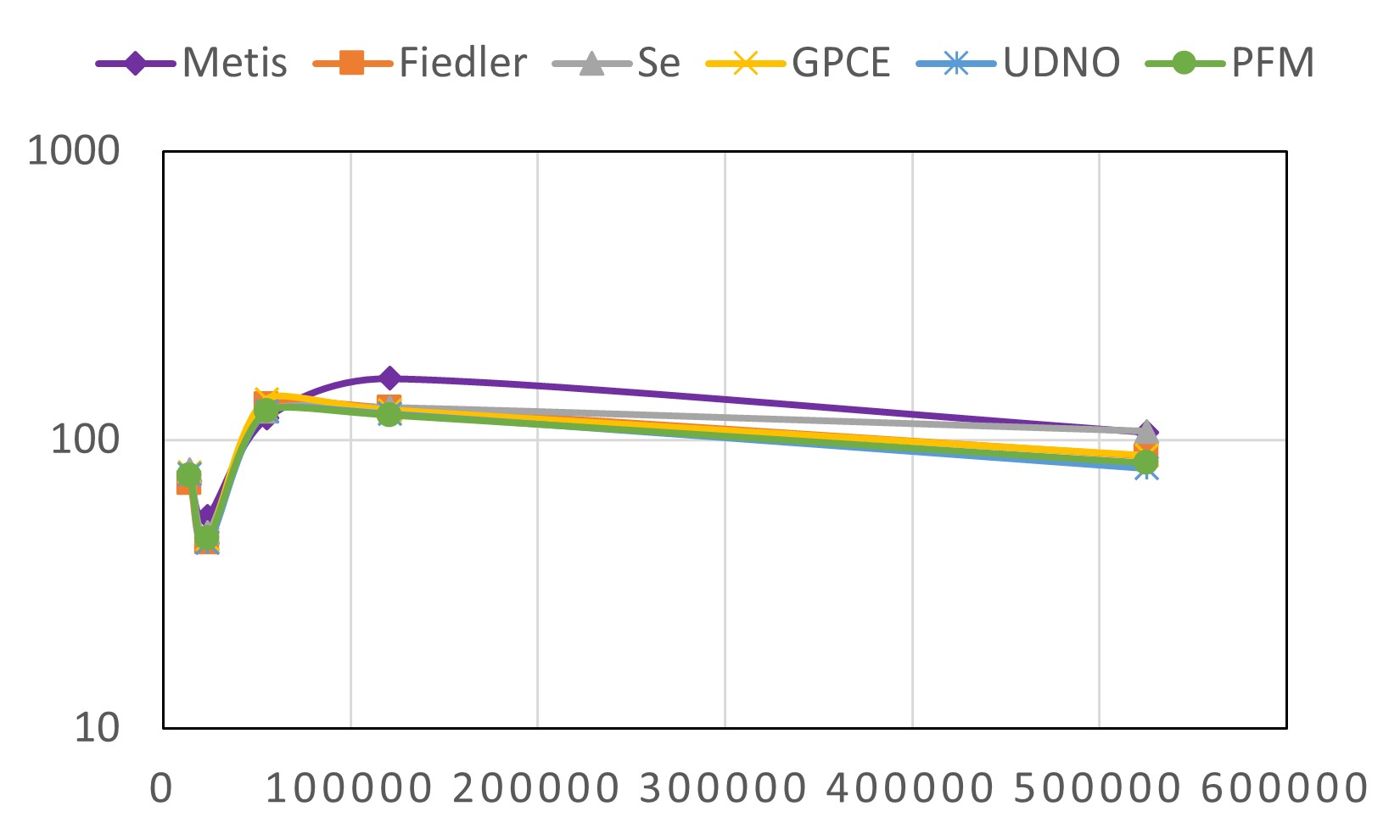}
}
\subfloat[LU Time (Seconds)]{
\includegraphics[width=0.33\textwidth]{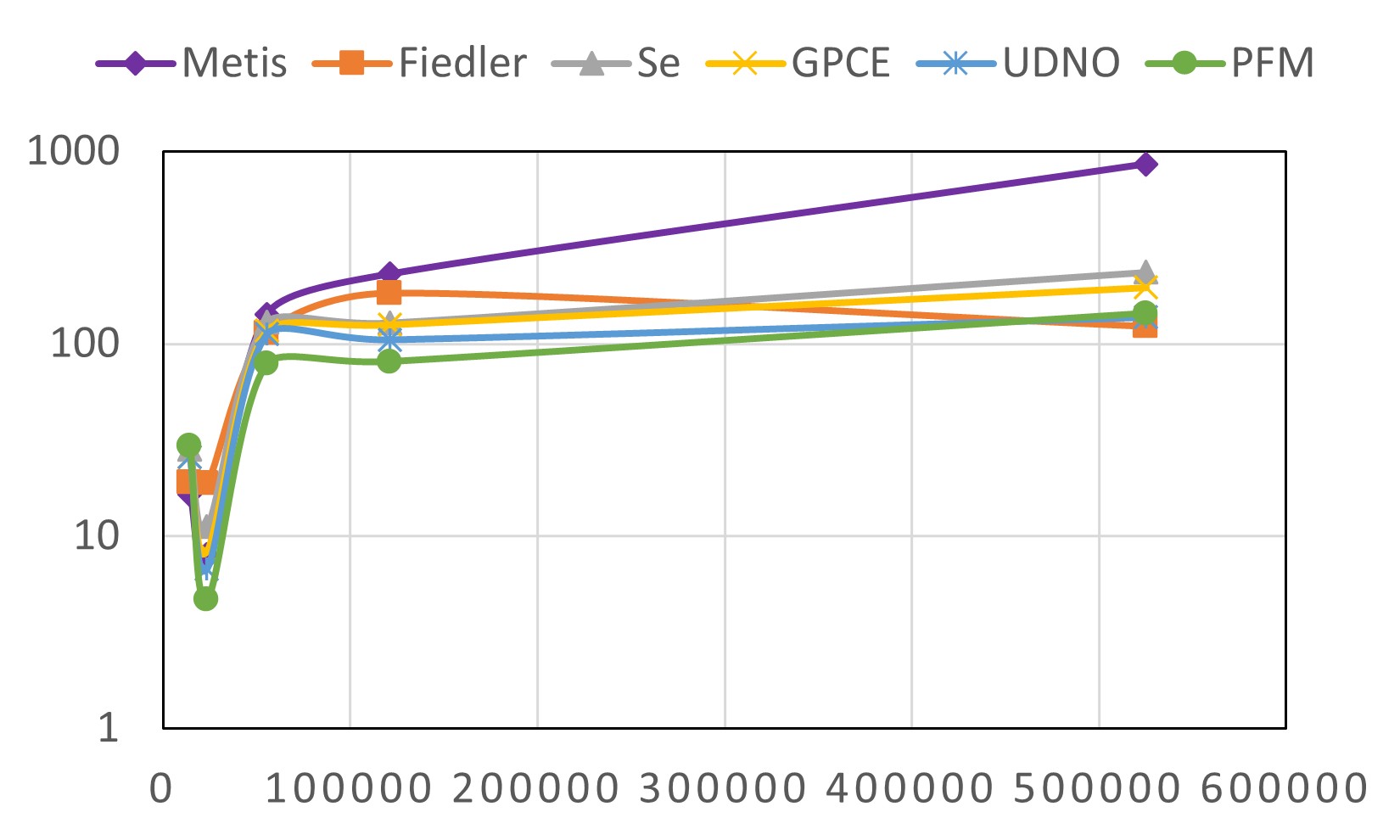}
}
\subfloat[Ordering Time (Seconds)]{
\includegraphics[width=0.33\textwidth]{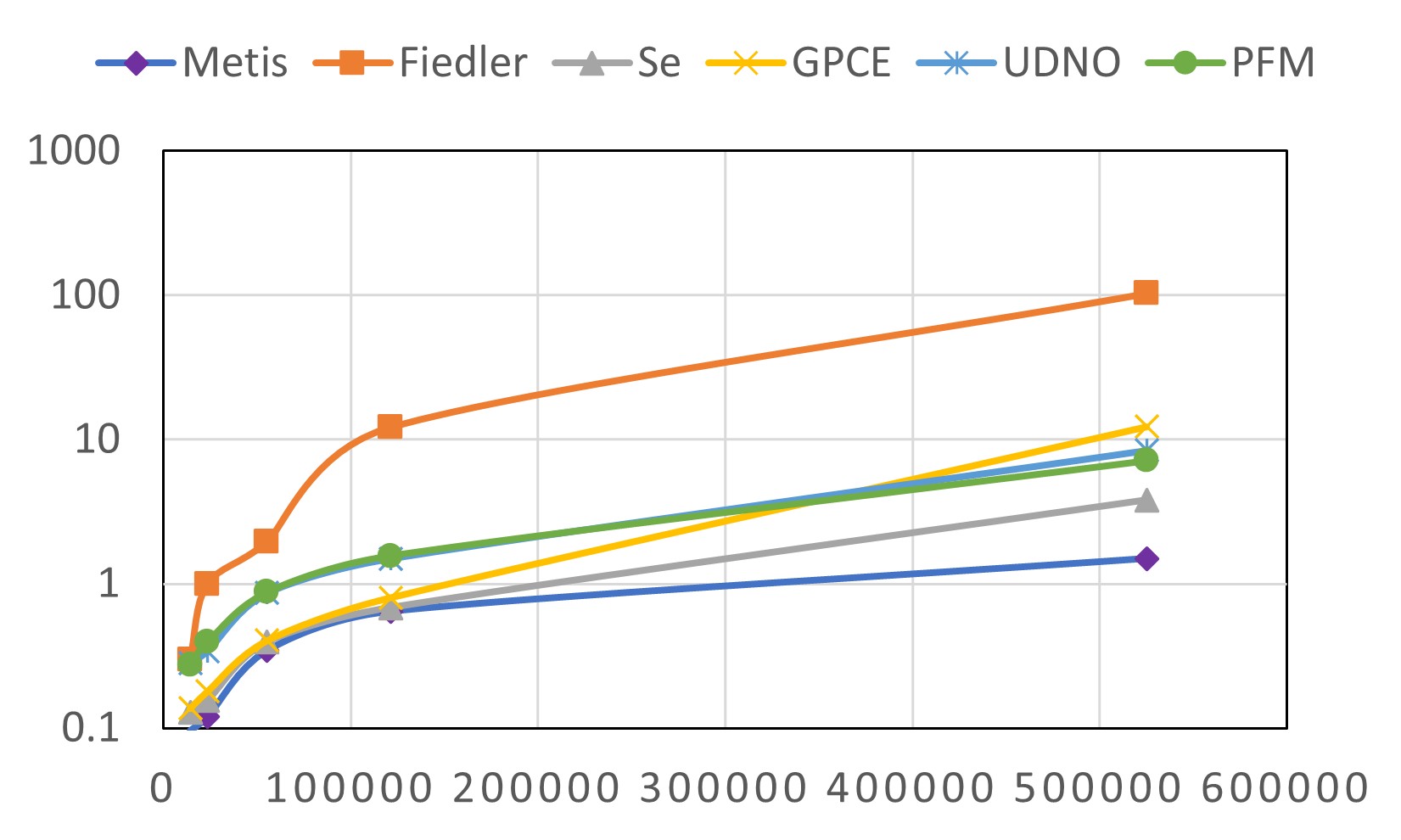}
}
\caption{Performance Variation along with Matrix Size for Matrix reordering Methods on SuiteSparse.}
\label{fig:scal}
\end{figure*}
\subsection{Effectiveness Study}

The fill-ins and LU factorization times obtained under different ordering methods on benchmark test set SuiteSparse from different scientific computing problems are shown in Table~\ref{tab:effect}. 

PFM achieves the best fill-in ratios and the least LU factorization time on most benchmark test matrices in Table~\ref{tab:effect}, such as matrices from structural problem, thermal problem and 2D3D problem. Thus it obtains the state-of-the-art performance on the whole test set compared with both graph theoretical methods and deep learning methods. Compared with existing state-of-the-art deep learning based reordering method UDNO, the fill-in ratio and LU factorization time are reduced by 0.15 and 9 seconds respectively.

It is worth noting that the reordering network of PFM is trained on a mixed matrix set with some 2D3D matrices. Thus it is reasonable that the fill-ins and factorization times on 2D3D matrices reordered by PFM are lower than those of other methods. For SP and TP matrices, PFM also obtains the best results among baselines in Table~\ref{tab:effect}. This indicates the generalization ability of PFM is improved compared with other deep learning methods. On some subsets like CFD and MRP, both fill-ins and factorization time are a little higher. The largest fill-in ratio gap is about 7.5\% between PFM and Metis on other kinds of matrices. On these subsets, the other deep learning methods like UDNO also fail. It indicates there is improvement space for its generalization ability to different scientific computing problems.

Compared with graph theoretical methods in Table~\ref{tab:effect}, PFM achieves the overwhelming performance in terms of both metrics, i.e. fill-in ratio and LU factorization time. The performance improvement is better explained by two possible reasons. One is the deep learning technique mentioned above helps gain some generalization ability. The other is the optimization objective function in the form of \(l_1\) norm, which is a well behaved and theoretically guaranteed approximation to the number of fill-ins.

PFM is consistently better than other deep learning baselines. Compared with GPCE, its advantage lies in the Multi-grid GNN and the proximal fill-in optimization. Compared with UDNO, their major difference lies in the objective function. UDNO optimizes an expected envelope like loss function and PFM focuses on the approximate fill-in minimization with \(l_1\) norm. A detailed analysis will be in the ablation study.

\begin{table}[!htbp]
\centering

\footnotesize
\begin{tabular}{|c|ccc|}
\hline
  & SP& CFD& SP+CFD\\
   \cline{2-4}
 \hline
\textbf{$S_e$} & 	55.66 &61.62&58.64\\
\hline
\textbf{randinit+MgGNN+FactLoss}&123.1&	-&-\\
\hline
\(S_e\)\textbf{+MgGNN+PCE}&51.48 	 	&59.80  &	54.46   \\
\(S_e\)\textbf{+MgGNN+UDNO}&46.67 &	59.57 &	53.12 \\
\hline
\(S_e\)+GUnet+PFM&71.55&107.58&89.56\\
\hline
\(S_e\)\textbf{+MgGNN+FactLoss}&45.32&58.30&51.81\\ 	 

\hline
\end{tabular}
\caption{Ablation Study of PFM.}\label{tab:ablationloss}
\end{table}

\subsection{Ablation Study}
PFM is mainly composed of three parts: the spectral embedding \(S_e\), graph node encoder with Multi-grid GNN (MgGNN) and the factorization enhanced loss (FactLoss). Here \(S_e\)+MgGNN+FactLoss means the proposed method PMF in Table~\ref{tab:ablationloss}. We replace its spectral embedding with random initialization and obtain the results of the model in the second rows. To explore the role of the loss function, we replace the factorization loss with pairwise cross entropy and UDNO's loss function, their results are shown in the third and fourth rows. The role of the network architecture is also explored in the fifth rows by replacing the MgGNN with GraphUnet.

The performance comparison between the second and last rows shows that the spectral embedding lies at the heart of the whole framework PFM. Without spectral embedding, the model achieves the worst result in the table. The performance difference between the last two rows indicates the network architecture also plays a significant role in the performance gain of PFM. Compared with the middle two rows, the last row achieves the best performance due to the proximal fill-in minimization framework.

\subsection{Scalability Analysis}
According to this matrix size distribution, we nearly uniformly divide the $148$ test matrices into five groups. For each group, the average performance of each ordering method is derived in terms of fill-in ratio, LU factorization time and ordering time and they are plotted in Figure~\ref{fig:scal}. We do not consider AMD and Natural methods due to their large performance differences from other methods.

Fill-in ratios in Figure~\ref{fig:scal}(a) change slowly along with the matrix size, especially when the matrix size is larger than $20$ thousand. This is because the fill-in ratio is calibrated by the number of nonzero elements, which is related to the matrix size. Without matrix size normalization, the LU factorization and ordering time will actually reflect the scalability of these ordering approaches. Both the LU time of Metis in Figure~\ref{fig:scal}(b) and the ordering time of Fiedler in Figure~\ref{fig:scal}(c) are out of control when the matrix size is larger than $100$ thousand. As we expected, the graph theoretical methods, i.e. Metis and Fiedler, are not so scalable as the deep learning method, such as $S_e$, GPCE, UDNO and PFM. Due to their excellent scalability, it is promising to further study deep learning methods for matrix reordering. Ordering time curve agrees with the time complexity in Table~\ref{tab:timecomplexity}.

Spectral embedding $S_e$ is trained on matrices with size less than $10$ thousands while both GPCE, UDNO and PFM are trained on matrices less than $1$ thousand. Though a large gap between training and test matrix size distribution, most deep learning methods achieve the comparable performances with those excellent graph theoretical algorithms. 
A better understanding of the generalization ability is that sparse patterns on smaller training matrices memorized by network parameters are applicable to larger test matrices.

\section{Conclusion}

We propose a GNN based Proximal Fill-in Minimization (PFM) framework that minimizes fill-ins during sparse matrix factorization by introducing $l_1$ norm minimization of the Cholesky factor. In the differentiable matrix ordering layer, reparameterization transforms the predicted scores from the reordering network into a differentiable permutation matrix. A factorization-enhanced loss is designed, and ADMM is employed to jointly optimize the factor $L$ and the network parameters $\theta$, ensuring sparsity of $L$ while improving the ordering. By addressing the non-differentiability of discrete reordering operations and fill-in estimation, the proposed method enables end-to-end optimization and consistently outperforms traditional graph-theoretic approaches and state-of-the-art deep learning techniques.

\appendix

\section{Architecture of the Graph Node Encoder}

\begin{figure}[!htbp]
    \centering
    \includegraphics[width=1.0\linewidth]{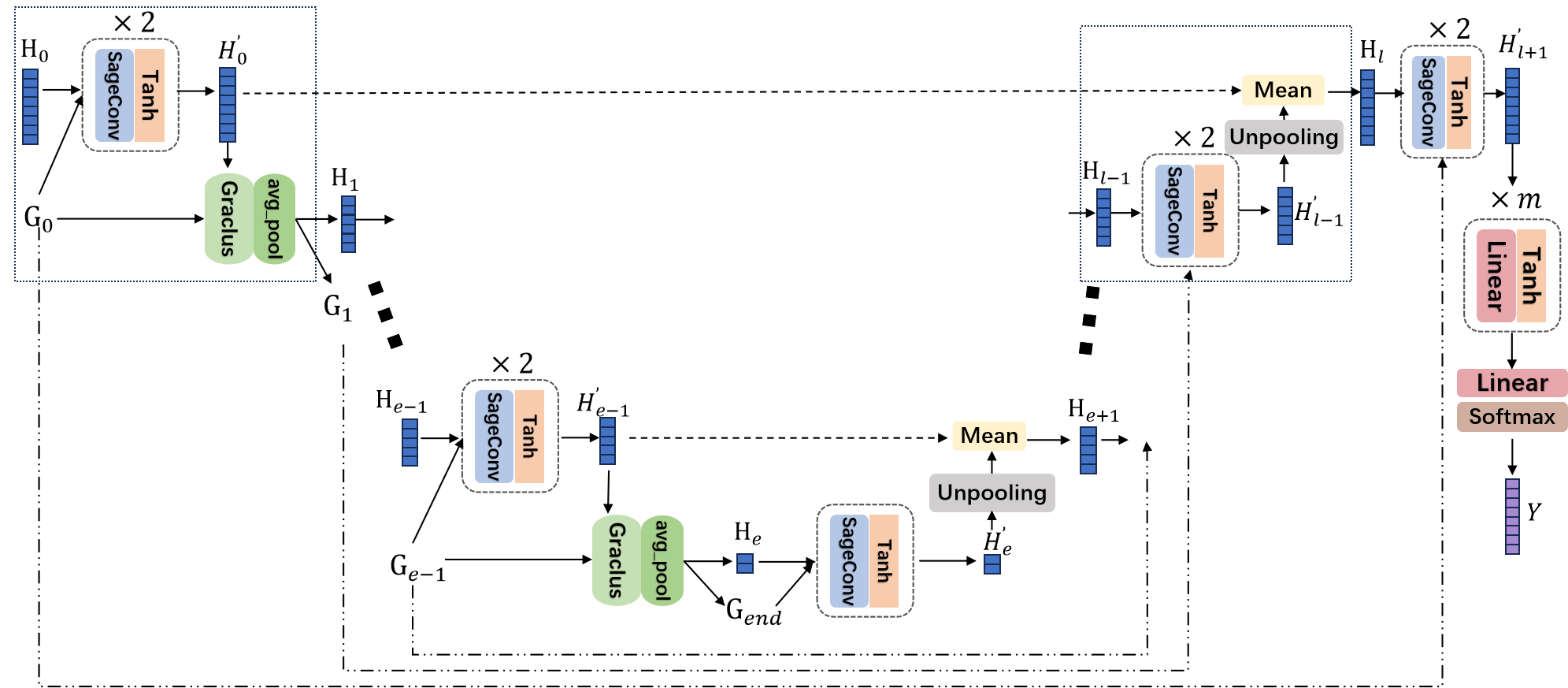}
    \caption{Architecture of the Graph Node Encoder}
    \vspace{10pt}
    \label{Network}
\end{figure}

This Graph Node Encoder network alleviates the locality limitation of naive GNNs through a hierarchical coarsening and refinement process.
Starting with the initial embedding $H_0=X_G$ and the original graph $G_0=G$, the module consists mainly of the pooling and unpooling stage, as shown in Figure~\ref{Network}. 

In the pooling stage, the node representations $H_c'\in R^{}$ are updated through two SAGEConv layers on the current coarse graph $G_{c}$ in Eq.(\ref{eq:nodeembedcoarse}) before each pooling operation is performed by the graph clustering method Graclus~\citep{graclus} to obtain a coarse graph $G_{c+1}=(V_{c+1},E_{c+1})$. 
Each clustering information $E_c$ and the updated node embeddings \( H_c' \) are pushed into stacks $S_E$ and $S_X$, respectively. The pooling stage continues until only two nodes remain in the coarsened graph. For the coarsest graph $G_{\text{end}}$, only one SAGEConv layer is used to update the node embedding, and no more pooling is needed. 
\begin{equation}
\label{eq:nodeembedcoarse}
H_c'=\text{Tanh}(\text{SAGEConv}(\text{Tanh}(\text{SAGEConv}(H_{c},E_{c}))))
\end{equation}

In the unpooling stage, the node embedding \( H_{l-1}' \) undergoes the unpooling operation to obtain \( H_{l-1}'[S_E.pop()] \). The node embedding \( H_l \) is then interpolated using \( H_{l-1}'[S_E.pop()] \) and \( S_X.pop() \), as shown in Eq.(\ref{eq:interplate}). Here, \( H_{l-1}' \) and \( S_X.pop() \) represent node features from the current and coarser graphs, respectively, while \( S_E.pop() \) and \( S_X.pop() \) are popped from the stacks \( S_E \) and \( S_X \), respectively.
Then node embedding $H_l$ is further smoothed through two SAGEConv layers similar to Eq.(\ref{eq:nodeembedcoarse}), resulting in the updated embeddings \( H_l' \). 
\begin{equation}
    \label{eq:interplate}
    H_l=(H_{l-1}'[S_E.pop()]+S_X.pop())/2
\end{equation}

After stacks $S_E$ and $S_X$ are empty, node embedding is completely updated. Finally, several linear layers are applied to output the ranking scores $Y=f_{\theta}(X_G)\in R^{n \times 1}$.

In this architecture, the initial node feature matrix $H_c \in \mathbb{R}^{n \times 1}$ contains one input feature per node. The first SAGEConv layer maps the node features from 1 dimension to a hidden dimension of $16$, while all subsequent SAGEConv layers maintain the $16$-dimensional representation. In the final stage, four linear layers are applied: the first three layers output $16$-dimensional node features, and the last layer outputs a scalar score for each node, yielding $Y \in \mathbb{R}^{n \times 1}$.

\section{Gumbel Sinkhorn Iteration}
Algorithm~\ref{alg:sinkhorn} provides the pseudocode for the Gumbel Sinkhorn Iteration Method.
\begin{algorithm}[htbp]
\caption{Gumbel Sinkhorn Iteration Method}
\label{alg:sinkhorn}
\begin{algorithmic}[1]
\REQUIRE Initial matrix $\hat{P}$, temperature parameter $\tau$.
\ENSURE Permutation matrix $P_{\theta}$.
\STATE /* Generate Gumbel noise */
\STATE $U = \text{rand}(\hat{P}.\text{shape})$;
\STATE $\epsilon_G = -\log(\epsilon - \log(U + \epsilon))$; 

\STATE /* Add Gumbel noise */
\STATE $\text{logP} = \log(\hat{P}) + \epsilon_G$;

\STATE /* Control determinism through temperature */
\STATE $\text{logP} = \frac{\text{logP}}{\tau}$; 

\STATE /* Sinkhorn normalization */
\FOR{$i = 1 \dots n_{\text{iters}}$}
    \STATE $\text{logP} = \text{logP} - \text{logsumexp}(\text{logP}, \text{dim}=0)$; 
    \STATE $\text{logP} = \text{logP} - \text{logsumexp}(\text{logP}, \text{dim}=1)$; 
\ENDFOR

\STATE $P_{\theta} = \exp(\text{logP})$; 
\RETURN $P_{\theta}$
\end{algorithmic}
\end{algorithm}

\section{Acknowledgments}
The work was supported by National Key R\&D Program of China (Grant No. 2021YFB0300203).
\bibliography{aaai2026}

\end{document}